\documentclass[10pt,twocolumn,letterpaper]{article}

\usepackage{cvpr}
\usepackage{times}
\usepackage{epsfig}
\usepackage{graphicx}
\usepackage{amsmath}
\usepackage{amssymb}
\usepackage{bm}
\DeclareMathOperator*{\argmax}{arg\,max}  

\usepackage[pagebackref=true,breaklinks=true,letterpaper=true,colorlinks,bookmarks=false]{hyperref}

\cvprfinalcopy 
\def\cvprPaperID{2} 

\begin{document}

\title{CPARR: Category-based Proposal Analysis for Referring Relationships}


\author{Chuanzi He \quad \quad Haidong Zhu \quad \quad Jiyang Gao \quad \quad Kan Chen \quad \quad Ram Nevatia\\
University of Southern California\\
{\tt\small {\{chuanzih|haidongz|jiyangga|kanchen|nevatia\}@usc.edu}} 
}

\maketitle

\begin{abstract}
The task of referring relationships is to localize subject and object entities in an image satisfying a relationship query, which is given in the form of \texttt{<subject, predicate, object>}. This requires simultaneous localization of the subject and object entities in a specified relationship. We introduce a simple yet effective proposal-based method for referring relationships. Different from the existing methods such as SSAS, our method can generate a high-resolution result while reducing its complexity and ambiguity. Our method is composed of two modules: a category-based proposal generation module to select the proposals related to the entities and a predicate analysis module to score the compatibility of pairs of selected proposals. We show state-of-the-art performance on the referring relationship task on two public datasets: Visual Relationship Detection and Visual Genome.
\end{abstract}

\section{Introduction}
\begin{figure}[t]
\begin{center}
\includegraphics[width=3.28in]{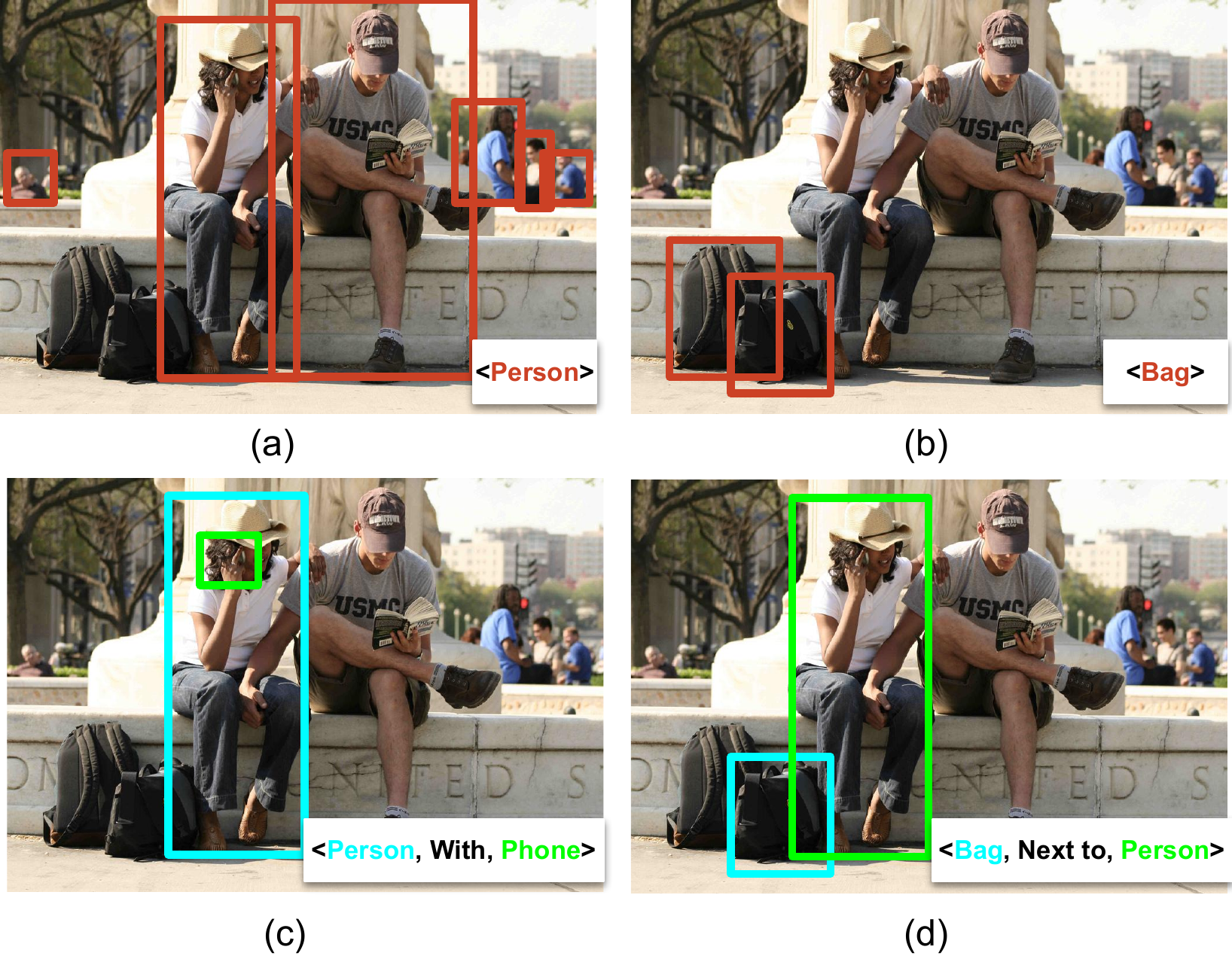}
\end{center}
   \caption{In a complex scene, referring relationships helps to localize target entities by their relationships with others. When querying for ``person" and ``bag", (a) and (b) give multiple instances of the same entity. If we want to localize a specific target, such as the person making a phone call, or the bag close to that person, querying with the relationship triplets \texttt{<person, with, phone>} in (c) and \texttt{<bag, next to, person>} in (d) helps by localizing both the subject and object entities.}
\label{fig:1}
\end{figure}

Localizing the entity in an image that is specified by a textual query, which can refer to both a single noun and its properties, such as ``a large, red sedan", has been an active area of research over the last few years \cite{chen2017query,hu2016natural,rohrbach2016grounding}. There has been recent work \cite{krishna2018referring} in including relationships between two objects in the queries, which have been called  \textit{referring relationships}. 
Such relationships are useful for various applications including image retrieval and visual question answering. Fig.~\ref{fig:1} shows examples where queries, ``person with phone" and ``bag next to person", help in differentiating a person and a bag from others in the same scene.

We consider a query to be in the form of \texttt{<subject, predicate, object>}. 
The problem of grounding entities in a relationship is more challenging than noun phrase grounding, as it subsumes the task of single object grounding and imposes the requirement of satisfying a relationship between a pair of objects. Modeling predicates is difficult due to the imprecise definition of relations. For example, in ``next to" and ``near",  the expectations of distances between entities may depend on the types of entities involved; distances are not the same in \texttt{<bag, next to, person>} and \texttt{<car, next to, building>}. 
Different from the tasks such as \textit{visual relationship detection} \cite{lu2016visual,yu2017visual} and \textit{scene graph generation} \cite{xu2017scenegraph}, which also explore the detection of \texttt{<subject, predicate, object>} triples, the task of referring relationships focuses on the relationship between the specific subject and object pairs given in the query. Methods in visual relationship detection and scene graph generation attempt to find all relationships in an image; so, presumably, the queried triples will also be in the output set, but it may possibly be discarded due to the potential large number of relationships. The detection model may also focus on more common relationships such as ``person standing" than ones with lower frequency due to the imbalance of relationships in the training set.
An existing state-of-the-art method, SSAS \cite{krishna2018referring}, aims to avoid the difficulty of variations in the appearance of subject-object pairs by generating two attention maps to influence each other by shifts, but the accuracy of the inferred bounding boxes suffers due to the low resolution of attention maps.

In this paper, we introduce a proposal-based method which is composed of two steps: first using a category-based proposal generating module to localize and select related candidate proposals based on their categories for subject and object entities and then applying a predicate analysis module to identify proposal pairs satisfying the queried predicate. By decoupling the proposal generation with the predicate analysis, the network can first pick out highly related entities to reduce both the complexity and ambiguity for predicate prediction and then analyze the relationships between selected proposals.
We call our complete system as CPARR for ``\textbf{C}ategory-based \textbf{P}roposal \textbf{A}nalysis for \textbf{R}eferring \textbf{R}elationships". With category-based proposals for related candidates and specified predicate analysis, we show state-of-the-art performance on the public datasets for referring relationships with different evaluation metrics. 

In summary, our contributions are two-fold: 1) a category-based proposal generator to select related candidates and tackle the challenge of accurate localization; 2) a predicate analysis network trained with selected proposals to model the role of the predicate in disambiguating object pairs.
In the following, we first introduce related work in Sec.~\ref{sec: related work}, then we provide details of CPARR in Sec.~\ref{sec: method}. Lastly, we present the evaluation and comparison with baseline methods in Sec.~\ref{sec: exp}, followed by conclusions in Sec.~\ref{sec:conclusion}.

\begin{figure*}
\begin{center}
\includegraphics[width=50em]{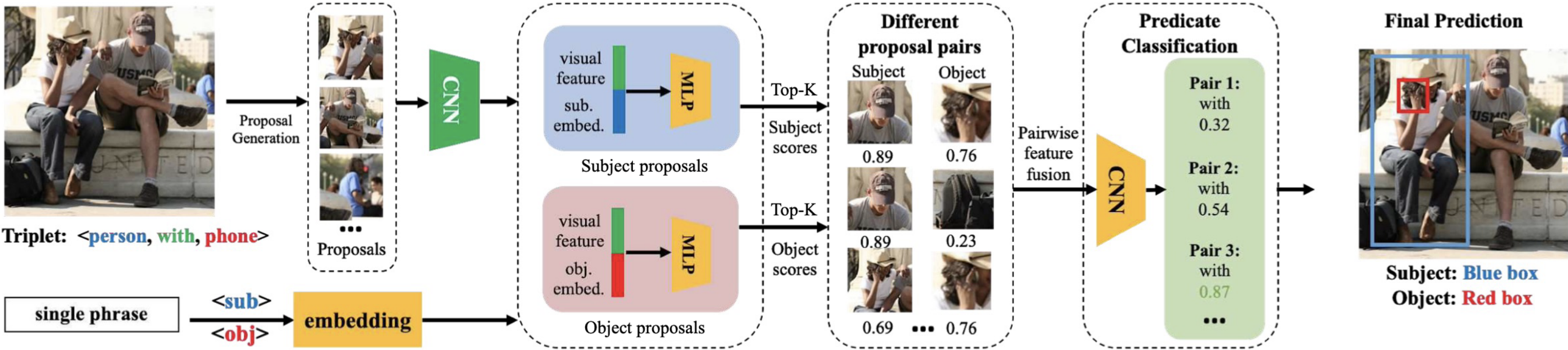}
\caption{The framework of CPARR. The category-based proposal generating module tackles the case when the input query only indicates one entity, i.e., the subject or the object entity with its phrase embedding result. The predicate analysis module considers the whole relationship phrase and disambiguates subjects and objects proposed by the category-based proposal generating module in this relationship.
}
\label{fig:framework}
\end{center}
\end{figure*}

\section{Related Work}\label{sec: related work}
There has been limited work directly on the referring relationships task. However, the tasks of scene graph generation, visual relationship detection, human-object interaction and phrase grounding have some relations; we briefly summarize them in the following.

\textbf{Scene Graph Generation:} To find the relationship pairs, some researchers generate scene graphs \cite{chen2019knowledge,yikangfactorizable,li2017scene,xu2017scenegraph,Yang_2018_ECCV,zellers2018motif} for the dense relationships reconstruction in the image. A scene graph represents entities and all the relationships in a graph where the nodes represent entities and the edges represent the relationship between the nodes. Xu \emph{et al.}~\cite{xu2017scenegraph} provides an end-to-end solution built with standard RNNs and iterative message passing for prediction refinements. Neural Motif \cite{zellers2018motif} observes statistics of relationships labels and utilizes motifs, regularly appearing substructures in scene graphs. Factorizable Net \cite{yikangfactorizable} replaces numerous relationship representations of the scene graph with fewer subgraphs and object features to reduce the computation. 

\textbf{Visual Relationship Detection:} Finding all the existing triplet relationships \texttt{<subject, predicate, object>} in a scene is also explored in the Visual Relationship Detection (VRD) task \cite{dai2017detecting,li2017vip,liang2018visual,lu2016visual,peyre2019detecting,yu2017visual,yin2018zoom}. Yu \emph{et al.}~\cite{yu2017visual} leverages external datasets and distills knowledge for triplet training and inference. Shuffle-then-Assemble \cite{yang2018shuffle} applies unsupervised domain transfer to learn an object-agnostic relationship feature. Zoom-Net \cite{yin2018zoom} proposes spatially and contextually pooling operations to improve feature interaction between proposals. Different from referring relationships, it is not easy to find out the subject and object entities in VRD due to the exponential number and its long-tailed distribution of entity types and their combinations, which might also result in the required entities being discarded due to the low interest.

\textbf{Human Object Interaction:} Human Object Interaction focuses on detecting and recognizing how human in the image interacts with the surrounding objects \cite{chao2018learning,gao2018ican,gkioxari2018detecting,li2019transferable,qi2018learning,wan2019pose,xu2019learning}. ICAN \cite{gao2018ican} uses the appearance of an entity to learn the highlight informative regions. Xu \emph{et al.} \cite{xu2019learning} implements knowledge graphs for modeling the dependencies of the verbs and objects. Compared with referring relationships, HOI only has one subject class, while both subjects and objects in referring relationships tasks can be human or objects. Also, compared with HOI, relationships described in referring relationships are much more varied.

\textbf{Phrase Grounding and Referring Expression:} Phrase grounding and referring expression apply the visual and language modalities to solve the problem of localizing entities for specific queries \cite{chen2018knowledge,chen2018msrc,chen2017query,galleguillos2008object,hu2016natural,krishna2018referring,laberge1997shifting,liu2019knowledge,Plummer2017PhraseLA,rohrbach2016grounding,sadhu2019zero,wang2016structured,yu2018mattnet}.
SSAS \cite{krishna2018referring} uses attention maps for localization. However, due to the low resolution of the generated attention map (14 $\times$ 14), the inferred bounding boxes are less accurate. 
Chen \emph{et al.}~\cite{chen2017query} introduces the regression mechanism and reinforcement learning techniques to improve the grounding performance. 
MAttNet \cite{yu2018mattnet} uses modular components including subject, location and relationships, to adaptively process the expression contents. Compositional Modular Network \cite{hu2017modeling} decomposes the task into modular networks handling language parsing, localization and pairwise relationships. Compared with phrase grounding and referring expression, the referring relationships task focuses on finding the correct entities based on the relationship, where strong hints such as location do not exist. 

\section{Method}\label{sec: method}
Our goal is to infer the location of the queried subject and object when given an image $I$ and a relationship query \texttt{q=<S, P, O>}, where S, P and O represent the categories of the subject, predicate and object. In this section, we will first formulate the problem and then introduce the category-based proposal generating and predicate analysis module respectively, followed by the implementation details.

\subsection{Problem Formulation}
We take an image, $I$, and a triplet query \texttt{q=<S, P, O>} as the input of the network with parameter $\theta_{M}$. To obtain the location of the subject, $y_{s}$, and object, $y_{o}$, conditioned on the given query \texttt{q} respectively, we express this inference task as a probabilistic problem which is shown as follows:
\begin{equation}
\begin{aligned}
\label{eq: prob formulation}
P(y_s, y_o | \langle S, P, O \rangle) =\argmax_{y_s, y_o, \Theta_{M}}&P(y_s | \langle S \rangle)\cdot P(y_o | \langle O \rangle) \cdot \\
 &P( P|(y_s, y_o))
\end{aligned}
\end{equation}

\subsection{CPARR}

Our method solves the localization precision challenge in two main steps: first, it finds related candidates for subjects and objects by selecting them independently using their descriptions in the query and then pick out pairs that best satisfy the given predicate.  Fig.~\ref{fig:framework} provides an overview of our proposed framework. Object proposals and their features are generated from an image and then passed to two category-based proposal generating modules, one for the subject and one for the object. These two modules have an identical architecture but do not share weights. After proposals are ranked, the predicate analysis module takes pairs from the top-ranking outputs of the two category-based proposal generating modules and evaluates them for consistency with the given predicate which results in the selection of subject and object entities and their locations. In this subsection, we first introduce the category-based proposal and predicate analysis and then describe how these two separate parts are combined to make the final inference.

\subsubsection{Category-based Proposals}\label{spg}

To generate category-based proposals, an entity is localized by a bounding box with a noun phrase from the query
regardless of its relationship with other entities. 
We use two independent category-based proposal generating modules $M_{sub}$ and $M_{obj}$ to regress and predict probability scores for subject and object entities respectively.  

We extract a set of $N$ candidate proposals $\{B_i\}_{i=1}^N$ from the image $I$ by using a Region Proposal Network \cite{ren2015faster} as initial bounding boxes and extract feature vectors $\{\bm{f}_i\}_{i=1}^N$ corresponding to each region. We represent the 5-dimension spatial feature of $B_i$, which is $[\frac{x_{min}}{w_I}, \frac{y_{min}}{h_I}, \frac{x_{max}}{w_I}, \frac{y_{max}}{h_I}, \frac{ Area_{B_i}}{Area_{I}}]$, as $\bm{s}_i$.
The full representation of a proposal, $\bm{v}_i$, is the concatenation of visual feature $\bm{f}_i$ and spatial feature $\bm{s}_i$. The input of the network to generate category-based proposals is the concatenation of visual features and phrase embedding vectors of the proposals. The network first transforms the visual feature $\bm{f}_i$ and the embedding vector of the subject or object phrase, $\bm{e}_p$ into 
a multimodal space following 
\begin{equation}
\bm{m}_i = \phi(\mathbf{W}_m(\bm{v}_i || \bm{e}_p) + \mathbf{b}_m)
\end{equation}
where the multimodal feature $\bm{m}_i\in\mathbb{R}^{128}$ aims to align the visual appearance and the semantics so that the predicted probabilities are conditioned on both the proposal's visual appearance and the subject/object category. $\mathbf{W}_m\in\mathbb{R}^{d_i\times 128}$ is the projection weight and $\mathbf{b}_m\in\mathbb{R}^{128}$ is the bias. 
$||$ represents the concatenation operator and $\phi(.)$ is the non-linear activation function. After a multi-layer perceptron layer network, $M_{sub}$ and $M_{obj}$ give the multimodal embedding of each candidate, $\bm{m}_i$, a confidence score $\bm{c}_i$ and provide regression offsets $\bm{t}_{i}$ to refine the initial bounding box.
The calculation of 4D regression parameters $\bm{t}_{i}$ is defined as $[(x-x_{a})/w_a, (y-y_{a})/h_a, log(w/w_{a}), log(h/h_a)]$, following \cite{ren2015faster}, where $x$ and $x_a$ are for the predicted box and anchor box respectively.

$M_{sub}$ and $M_{obj}$ have two objective functions, 1) $\mathcal{L}_{cls}$ for predicting the confidence of $B_i$ being the phrase embedding of the queried entity $\bm{e}_p$ and 2) $\mathcal{L}_{reg}$ showed in Eq.~\ref{eq: candidate reg loss} for regression offsets that adjust the initial boundaries of $B_i$ conditioned on the input query. We assume there can be more than one candidate overlapping with the groundtruth with an Intersection Over Union (IoU) larger than a threshold $\tau$ and consider all these candidates to be positive. The loss of classification objective function is measured by the sigmoid cross-entropy loss. 
The regression offsets calculate L1-smoothness regression loss between the positive candidates $\bm{t}_{i}^{p}\in\mathbb{R}^{4}$ and the groundtruth $\bm{t}_{i}^{q}\in\mathbb{R}^{4}$, where $f(.)$ is the smooth L1 loss function.  $N$ is the number of positive candidates $i*$ after regression offsets. 

\begin{equation}
\label{eq: candidate reg loss}
\mathcal { L } _ { r e g } \left(  \bm { t } _ { i } ^ p ,  \bm{t} _ { i } ^ q  \right) = \frac { 1 } { 4 N } \sum _ { i = 1 } ^ { N } \sum _ { j = 0 } ^ { 3 } f \left( \left| \bm { t } _ { i } ^ { p } [ j ] - \bm { t } _ { i } ^ { q } [ j ] \right| \right)
\end{equation}

We rank the candidates by confidence $\mathbf{c}_i$ to perform offset regression on the best proposals and feed the top-$K_{sub}$ and top-$K_{obj}$ proposals to the next module.

\subsubsection{Predicate Analysis} \label{sec:pred_cls_module}
The category-based proposals are to localize entities across different categories, while the disambiguation of subject and object entities depends on inter-object relationships, in particular, the predicate connecting a subject and an object. The predicate analysis module selects subject and object entities that participate in the same relationship query by evaluating the predicate category between a pair of proposals.

Following the category-based proposal generation, the input to the predicate analysis module is a pair of proposals, $B_i$ and $B_j$. The module $M_{pred}(B_i, B_j)$ outputs predicate confidence scores of $\{B_i, B_j\}$ under ${P+1}$  predicate categories, with $P$ being the total number of predicate categories plus one for the background class where the pair does not have any of the enumerated relationships. The network first concatenates visual features of $B_i$ and $B_j$, 
then compresses the dimension by a convolutional neural network, and finally outputs a score for verification.  We take the score corresponding to the predicate type in \texttt{q = <S, P, O>} as the probability of $Prob (P |B_i, B_j) $, representing $B_i$ and $B_j$ forming the queried relationship $P$. 

To recognize the relationship between two regions of the image, their appearance similarity, spatial connection, interaction with other regions all contribute to the recognition results. Therefore, there is a demand for effective proposal feature interaction to comprehensively exploit useful appearance, spatial and semantic interaction between the proposal pairs. In our method, instead of using one-dimensional feature vectors, we concatenate two $W \times H \times D$ spatial feature maps that come from 
ROI pooling \cite{ren2015faster} 
depth-wise to form a $W \times H \times (D \times 2)$ dimension input tensor. The consideration is that the multi-dimensional feature maps incorporate spatial information and contextual visual features. The subject candidate $B_i$ and object candidate $B_j$, which form the pair $\{B_i, B_j\}$,  come from $M_{sub}$ and $M_{obj}$ respectively. When constructing pairs, we take $K_{sub}$ subject proposal candidates and $K_{obj}$ object proposal candidates, forming $K_{sub} \times K_{obj}$  pairs for each query \texttt{q}.

The correct classification should only identify pairs with the positive subject and object candidate pairs as the known \texttt{<predicate>} category. The role of this module, classifying the presence of a predicate, requires constructing a training set with positive examples and two types of negative examples: 
i) $B_i$ or $B_j$ is not a correct proposal for the subject or object entity, and
ii) $B_i$ and $B_j$ do not form any relationships in the $P$ given categories.

\subsubsection{Combined Inference}
The model combines probabilities from the category-based proposals and predicate analysis for final inference following Eq.~\ref{eq: prob formulation}, and the candidate object and subject proposals for one query are selected as the ones which yield the highest probability. With the final $K_{sub} \times K_{obj}$ predicate classification scores, we select candidates with high weighted confidence on the category-based proposals and predicate verification as correct prediction. Note that if the predicate confidence is under a threshold $\tau_{pred}$, we set the weight of predicate confidence as 0 and solely use the category-based proposal score, because its predicate confidence could be low due to the inaccurate pairing candidates, which result in errors accumulated by the category-based proposals. 

\subsection{Implementation Details}
In this subsection, we present the implementation details of our method. We first introduce how the proposals and features for the two stages are generated, then show our network structure for category-based proposal generation and predicate analysis modules separately, and finally, we show our detailed information on training and testing.

\textbf{Proposal Generation:} We use a pretrained RPN \cite{ren2015faster} to generate initial candidate proposals. 
The RPN is initialized with the VGG16 \cite{simonyan2014very} pre-trained on ImageNet \cite{deng2009imagenet} and then trained on the datasets in the experiments. We set Non-Maximum Suppression (NMS) in RPN as 0.6 and generate $N=300$ proposals for each image after RPN to feed it into the category-based proposal generating network.

\textbf{Visual Features Extraction:} 
After the proposals are generated, we use a ResNet-50 \cite{he2016deep} pre-trained on ImageNet \cite{deng2009imagenet} followed by an average pooling layer \cite{chen2017implementation} to extract proposal features from bounding boxes. In the category-based proposal generation, proposal features are feature vectors from an average pooling layer. In the predicate analysis module, the feature maps from the ROI pooling layer are directly used as the input of visual features. 

\textbf{Phrase Embedding Generation:} For the subject and object phrases, we use the GloVe embedding algorithm \cite{pennington2014glove} to map a phrase to the 300-dim phrase embedding vector, which is then concatenated with the visual feature before sent for category-based proposals selecting.

\textbf{Network Architecture:} The category-based proposal generating network is a five-layer Multi-Layer Perceptron (MLP), where the first layer maps the concatenated visual and textual feature into a 128-D multimodal vector, followed by three 128-dim hidden layers and finally projects the vector to the 5-D output $\mathbf{c}_i || \mathbf{t}_i $. The predicate analysis module consists of 3 convolution layers with $3 \times 3$ kernels and one convolutional layer with $1 \times 1$ kernels. All nonlinear layers use ReLU activations.

\textbf{Training and Testing:}  
During training, We first train the RPN, then the two category-based proposal generating networks and finally the predicate analysis module. The outputs from the previous stages are used to train the next stage. We use the Adam optimizer \cite{kingma2014adam} with an initial learning rate of 0.0001. The maximum iteration is set to be 20000 on the category-based proposal generating module and 10000 on the predicate analysis module.
We adopt a multi-label training scheme in the category-based proposal generating module, so there could be multiple possible targets for classification. $\tau_{pred}$ is set to be 0.5. $ K_{sub}$ and $ K_{obj}$ are both set to be 5. 
For the predicate analysis module, the numbers of positive and negative examples are kept to be the same. We select positive and negative boxes from category-based proposals and train them with the Sigmoid cross-entropy loss. The predicate classification target of positive pairs $\{B_i,B_j\}$ is the predicate \texttt{<P>}, while target labels for negative pairs is the background predicate type $P+1$. 
For testing, we first apply an NMS on all proposals and then select the subject and object candidates with top-$K$ confidence, where $K$ is also set to be 5 empirically. The rate used for NMS is 0.5 in our experiments. The top $K$ confident subject and object proposals are selected as candidates for predicate analysis.

\section{Experiments}\label{sec: exp}

In this section, we provide results on benchmark datasets to show the performance of our model. For quantitative results, we compare with the four existing state-of-the-art methods on IoU score and recalls respectively. For qualitative results, we show some visualization results for subjects and objects entities with CPARR on the public datasets. 

\subsection{Datasets}
We evaluate our results on two popular visual relationship detection datasets with real scenes: VRD dataset \cite{lu2016visual} and Visual Genome \cite{krishna2017genome}. 

\textbf{VRD Dataset \cite{lu2016visual}:} The VRD dataset consists of 100 object types, 70 predicate types and 5000 images. In all, it contains 37,993 relationship annotations with 6,672 unique relationship types and 24.25 relations per entity category. 60.3\% of these relationships refer to ambiguous entities. 
Predicates are mainly from spatial, preposition, comparative, action, and verb types. We use the same dataset splits as in SSAS \cite{krishna2018referring} which consist of 4000 training samples and 1000 testing samples. 

\textbf{Visual Genome \cite{krishna2017genome}:}
Visual Genome is a dataset commonly used in scene graph generation and referring relationships evaluations. Following \cite{krishna2018referring}, we develop our results on version 1.4, which focuses on the top-100 frequent object categories and top-70 frequent predicate categories. We adopt the same subset of Visual Genome as used in SSAS \cite{krishna2018referring}, with 8560 images for the test set, 77257 images for the training and validation set.


\subsection{Evaluation Metrics}

For appropriate comparison with baseline methods, we first evaluate our results on the Mean IoU score. To compare with methods generating attention maps, we compute the IoU of heatmap and groundtruth following SSAS \cite{krishna2018referring}

\begin{equation*}
    IoU(Att, GT) = \frac{ \sum (\mathbb{I} (Att_{i} > \tau )\cap GT_{i})}{\sum (\mathbb{I} (Att_{i}> \tau) \cup GT_{i})}
\end{equation*}

where $Att_{i}$ and $GT_{i}$ denote the prediction and groundtruth for the $i$th cell in the heatmap. $\mathbb{I}$ converts prediction with IoU above the threshold $\tau$ as activated cells. To convert bounding boxes into heatmap masks, we first transformed the scale of bounding box coordinates down to the $L \times L$ heatmap size. The binary masks are obtained by setting regions within the bounding box as 1 and the outside as 0. To properly compare with previous methods \cite{krishna2018referring,lu2016visual}, L is set as 14. Note that our output bounding box is based on the original image size, we down-sample it to $L \times L$ for a fair comparison with SSAS \cite{krishna2018referring}.

To assess the precision of bounding boxes, we also evaluate the referring relationships using object detection metrics. 
In Visual Genome and VRD datasets, objects and relationship queries are not labeled exhaustively. Therefore we adopt \textit{Recall} of bounding boxes as a metric for localization evaluation. We directly apply the original results from VRD \cite{lu2016visual} for bounding boxes generation and directly use the code provided by SSAS \cite{krishna2018referring} to transform the heatmap into bounding boxes by first rescaling the heatmap to its original input image size, $224 \times 224$ and obtaining the bounding boxes by thresholding activations over $\tau$.

\subsection{Baselines}
We compare our method with four different baseline methods: CO \cite{galleguillos2008object}, SS \cite{laberge1997shifting}, VRD \cite{hu2017modeling,lu2016visual} and SSAS \cite{krishna2018referring}. SSAS \cite{krishna2018referring} is the present state-of-the-art method in referring relationships by using the attention map to iterate until the result converges, while SS \cite{laberge1997shifting} does not iterate. VRD \cite{lu2016visual} is the state-of-the-art method on the visual relationship detection problem by maximizing the similarity based on the embeddings for entities, which is the same as CO \cite{galleguillos2008object}, and finding extra relationship embeddings for classification.

\subsection{Discussion}
We compare our method with VRD \cite{lu2016visual} and SSAS \cite{krishna2018referring}, and highlight the differences and advantages of our method.

\textbf{Differences with VRD \cite{lu2016visual}:} VRD finds all triplet relationships in one image. It uses all proposal candidates from the detector and ranks all possible combinations in the image with their confidence. Due to a large number of possibilities,  only a certain number of top-scoring relationships are retained according to the evaluation. When applied to referring relationships, it is possible that queried relationships may not appear in the set of preserved relationships. In our method, the predicate analysis module interacts with the information only with the selected top-K candidates generated by the category-based proposal generation, which greatly reduces the complexity for the predicate analysis module by avoiding analysis on the likely irrelevant candidate proposals.

\textbf{Differences with SSAS:} SSAS  generates iterative attention maps to solve the problem of the referring relationships. It takes the whole image into consideration with high complexity, resulting in the final attention map to be low resolution. We decouple the task into two steps by generating category-based proposals first followed by relationship analysis to distinguish among a small set of the candidate. This both reduces the complexity and preserves the original resolution of the image.

\begin{table}[]
\begin{center}
    
{
\begin{tabular}{@{}c|cc|cc@{}}
\hline

& \multicolumn{2}{c|}{VRD Dataset}         & \multicolumn{2}{c}{Visual Genome} \\
\hline
 Method  & Subject & Object  & Subject & Object \\ 
\hline
CO \cite{galleguillos2008object}         & 0.347 &0.389 & 0.414 &0.490 \\
SS \cite{laberge1997shifting}           & 0.320 &0.371 & 0.399 &0.469 \\
VRD \cite{lu2016visual}        & 0.345 &0.387 & {\bf 0.471} &0.480 \\
SSAS \cite{krishna2018referring}           & 0.369 &0.410 & 0.421 &0.482 \\
\hline
CPARR          & {\bf 0.482} & {\bf 0.510}  & { 0.469}& {\bf 0.517} \\
\hline
\end{tabular}}
\end{center}
\caption{Mean IoU results on VRD dataset and Visual Genome dataset for subject and object entities.}
\label{table:iou-vrd}
\end{table}

\subsection{Method Variations} 
To evaluate the contributions of modules of CPARR, we define three variations: CPARR, CPARR-cp and CPARR-pa. 
CPARR is the complete system. 
CPARR-cp finds the result with the highest score obtained with the category-based proposals applied to subject and object entities independently; CPARR-pa finds the pair producing the highest predicate classification score for prediction, where the pairs are composed of top-scoring subject and object entities. 
Different from CPARR-pa, CPARR multiplies the predicate classification scores with the probabilities of the subject and object entities, while CPARR-pa only applies the predicate scores for final confidence prediction.  

\subsection{Quantitative Results}

We first compare CPARR with the baseline methods for IoU score, which is commonly used in referring relationships, and then compare CPARR with the state-of-the-art methods on the \textit{recall} metric since it can better reflect how good the methods are in finding correct subject and object entities. Lastly, we compare the performance of using top-K proposals and groundtruth, and show the result for finding the best proposal feature interaction.

\textbf{Mean IoU Score} For proper comparison with the existing baseline methods, we first show our mean IoU result on the Visual Relationship Detection dataset and Visual Genome dataset in Table \ref{table:iou-vrd}. 
Among the four baseline methods, the two existing state-of-the-art methods, VRD and SSAS, outperform the other two baseline methods, CO and SS. On the VRD dataset, CPARR shows significant improvements over the other four baseline methods for both the subject and object localizations, and for Visual Genome dataset, it has nearly the same accuracy on subjects and much better IoU result on objects.

\begin{table}[]
\begin{center}
    
\resizebox{.95\columnwidth}{!}{
\smallskip
\begin{tabular}{@{}c|ccc|ccc@{}}
\hline
& &subject&         & &object& \\
\hline
Method   & r@1   & r@5     & r@50    & r@1     & r@5   & r@50     \\
\hline
SSAS \cite{krishna2018referring}     & 0.215   & -      & -     & 0.242      & -        & -   \\
VRD \cite{lu2016visual}    & 0.315   & 0.388      & 0.391     & 0.349      & 0.403        & 0.404   \\
CPARR-cp & 0.450  & 0.663   & 0.864  & 0.496     & 0.666     & 0.842  \\
CPARR-pa & 0.384  & 0.586  & 0.864   & 0.401   & 0.609       & 0.842    \\
CPARR & \textbf{0.498} & \textbf{0.694} & \textbf{0.864} & \textbf{0.524} & \textbf{0.702} & \textbf{0.842} \\
\hline
\end{tabular}}
\end{center}
\caption{Recall on the VRD dataset. The results of subject and object localization are evaluated separately. CPARR-cp shows results of category-based proposal generating modules, where predicate is not involved. CPARR-pa shows localization with predicate classification scores. CPARR is the final result which combines CPARR-cp and CPARR-pa.}
\label{table:1-vrd-rec}
\end{table}

\begin{table}[]
\begin{center}
    
\resizebox{.95\columnwidth}{!}{
\smallskip
\begin{tabular}{@{}c|ccc|ccc@{}}
\hline
& &subject& & &object&  \\
\hline
Method   & r@1            & r@5            & r@50           & r@1            & r@5            & r@50           \\
\hline
SSAS \cite{krishna2018referring} & 0.230          &-               &-               & 0.291          &-               &- \\
CPARR-cp & 0.355          & 0.512          & 0.716          & 0.445          & 0.596          & 0.776          \\
CPARR-pa & 0.300          & 0.472          & 0.716          & 0.378          & 0.553          & 0.776          \\
CPARR & \textbf{0.375} & \textbf{0.527} & \textbf{0.716} & \textbf{0.464} & \textbf{0.613} & \textbf{0.776} \\

\hline
\end{tabular}}
\end{center}
\caption{Subject and Object Recall on the Visual Genome dataset.}
\label{table:2-vg-rec}
\end{table}

\begin{figure*}[h]
\begin{center}

\includegraphics[width=50em]{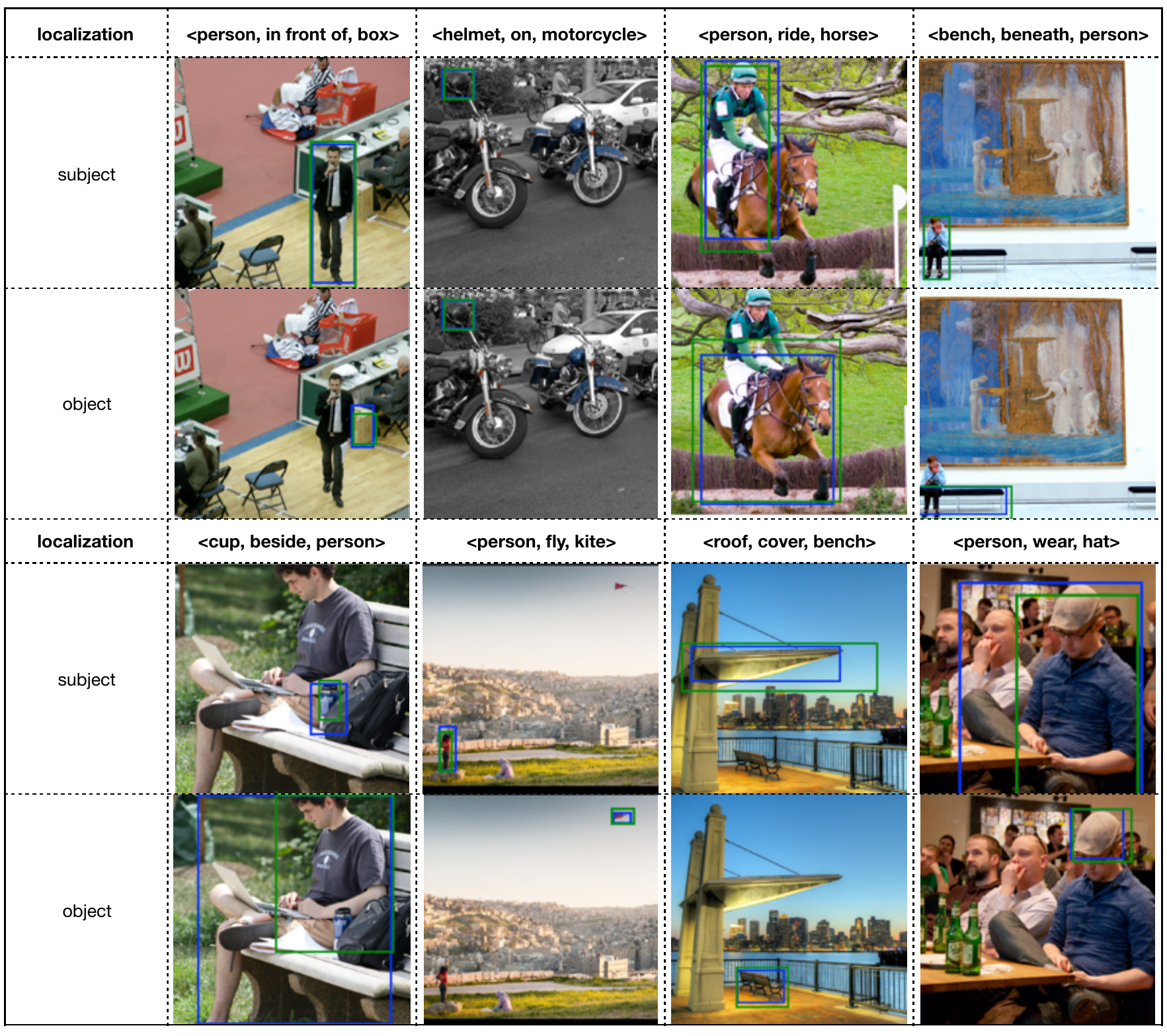}

\caption{Examples of CPARR results on VRD and Visual Genome dataset. The top rows are from VRD and the bottom ones are from Visual Genome. We visualize the groundtruth bounding box in blue and CPARR top-1 prediction in green. The captions above images are the \texttt{<subject, predicate, object>} triplet query. The top rows are localization results on \texttt{subject}. The bottom rows present \texttt{object} localization.}
\label{fig:demo}
\end{center}
\end{figure*} 

\textbf{Recall} Based on the IoU results, we select VRD and SSAS for object detection evaluation baselines method using the metric \textit{recall}. We get corresponding bounding boxes using the additional code\footnote{\url{https://github.com/StanfordVL/ReferringRelationships/blob/master/utils/visualization_utils.py}} provided by the authors for applying the bounding boxes for SSAS directly. Recall at top 5 and top 50 is not applicable since only one bounding box for subject and object can be obtained from the heatmap. 
Results for VRD\footnote{\url{https://github.com/Prof-Lu-Cewu/Visual-Relationship-Detection/blob/master/results}} are based on the detection results provided by the authors \cite{lu2016visual}. Table \ref{table:1-vrd-rec} and \ref{table:2-vg-rec} show our results for the recall on two datasets respectively.
Numbers in the table show recall of subject and object entities that have IoU with groundtruth of larger than 0.5 at three different ranks. 
Our method has superior performance over the two other baseline methods. Results also show that best results are obtained by combining detection and predicate scores (i.e. by CPARR). Note that VRD outperforming SSAS may be due to the inaccurate proposal results on the attention map with low resolution for SSAS. Better result comparing CPARR-cp with VRD may 
due to the category-based proposals being powerful enough to reduce the ambiguity compared with using all proposals.
This is also reflected in the result that CPARR-pa is not better than CPARR-cp, indicating that the prediction probabilities of the subject and object entities also play a significant role in the predicate analysis between subject and object pairs.

\textbf{Top-K Proposals Analysis:}
For further analysis of the performance of using top-K proposals for predicate analysis, we compare these results with those using groundtruth proposals in Table \ref{table: pred-cls-performance}. 
The predicate analysis module using GT proposals takes groundtruth subject and object locations as input, which demonstrates the performance of predicate analysis without the limitation of subject and object localization. 
The predicate analysis module using top-K proposals takes top-K proposals generated from the previous stage for training the classifier. K is set to be 5 in the experiment to show the final recall of predicate on the VRD dataset. The number of predicate categories is 70.
Results show that when training the predicate analysis module on Top-K proposals, the result is comparable to the model trained on groundtruth bounding boxes. When K is set to 5, the overall recall is comparably acceptable to provide the candidates including the correct proposals for training the predicate analysis. Instead of using all proposals, relationships generated from Top-K proposals can greatly reduce the complexity while still being sufficient to train a good predicate analysis model.

\begin{table}[]
\begin{center}
\begin{tabular}{ccc}
\hline
Method & r@1 & r@5 \\
\hline
GT proposals  & 0.7889  &    0.9609 \\
Top-K proposals &0.7365     &0.9168     \\
\hline

\end{tabular}
\end{center}
\caption{Evaluation of predicate analysis on the VRD dataset using groundtruth proposals and top-K proposals.}
\label{table: pred-cls-performance}
\end{table}

\textbf{Proposal Feature Interaction:} We compare different ways of proposal feature interaction and analyze their influence for predicate analysis and referring relationships on the VRD dataset in Table \ref{table:pred-cls-comparisons}. In the table, \textit{Vis Map} represents the ROI pooling feature, \texttt{<S,O>} represents phrase embedding $\bm{e}_p$ of subject and object categories, \textit{Vec} represents the visual feature vector $\bm{f}_i$ of the proposal, and \textit{Spatial} represents the 5D spatial feature $\bm{s}_i$ of the proposal. 
The four rows in the table represent predicate analysis module input settings as follows:
1) ROI-pooling feature as feature maps, 
2) the concatenation of feature vectors, 5D location vector(spatial feature), and phrase embedding feature \texttt{<S, O>}, 
3) a variant of case 2 but without phrase embedding features and
4) variant of case 2 but without spatial features as input.
We evaluate both the predicate classification accuracy and the recall result on object entities on the VRD dataset for CPARR-pa to show how it performs with different combinations of visual and location features.
From the results in Table~\ref{table:pred-cls-comparisons}, we make the following observations:

\begin{enumerate}
    \item[1)] For predicate verification, the ROI pooling feature maps, which preserve the multiple channel feature as well as its location, have the best performance over feature vectors representation and its variants.
    
    \item[2)] In all variants of feature vector-based pair representation, the concatenation with textual input of \texttt{<S, O>} and bounding box spatial information serve as effective hints for entity inference.
    
    \item[3)] Predicate classification score is higher with phrase embedding and spatial relation features, showing that spatial information and prior knowledge on subject and object combinations can provide useful content for predicting the predicate.
    
\end{enumerate}

\begin{table}[]
\begin{center}
\begin{tabular}{@{}l|llll@{}}
\hline
 & predicate & \multicolumn{2}{c}{CPARR-pa} \\ \hline
Feature Input   & r@1   & r@1     & r@5     \\ \hline       
Vis Map                                      & 0.7365   &   0.4012      & 0.6091        \\              
Vec + Spatial + \textless{}S,O\textgreater{} & 0.6680  & 0.3862  & 0.5918  \\              
Vec + Spatial                                & 0.6854  & 0.3335  & 0.5532  \\              
         
Vec       + \textless{}S,O\textgreater{}     & 0.6489  & 0.3742  & 0.6024  \\              
\hline
\end{tabular}
\end{center}
\caption{Recall of proposal visual and semantic feature combination on predicate classification for object entities on the VRD dataset. In PARR-pa, the r@50 results for all variants are 0.8642.}
\label{table:pred-cls-comparisons}
\end{table}

\subsection{Qualitative Results}\label{quality_result}
Besides quantitative comparison with existing baseline methods, we also visualize some examples from the VRD and visual genome datasets in Fig.~\ref{fig:demo}, where the detection results for subject and object entities are given separately. 
To focus on one example, in the \texttt{<person, wear, hat>} query, there are multiple ``person" entities given the query. In CPARR-cp, the top-5 ``hat" proposals result from all distribute around the hat on top of the second man to the right, giving strong hints to the person proposal which enclose the hat proposal at the top portion of the box, and correct the error of using the man left to the groundtruth as the result of ``person'', which actually has a higher score in CPARR-cp. 

\section{Conclusion}\label{sec:conclusion}

We introduce a proposal-based method with a category-based proposal generating module to pick out related candidates for subjects and objects separately to reduce the confusion and complexity of predicate prediction, and a predicate analysis module to further disambiguate subject and object entities to decide whether a subject-object pair belongs to a known predicate category. Our method has significantly higher accuracy than previous methods on multiple evaluation metrics on public datasets with real scenes for referring relationships. 

\section*{Acknowledgments}
This work was supported by the U.S. DARPA AIDA Program No. FA8750-18-2-0014. The views and conclusions contained in this document are those of the authors and should not be interpreted as representing the official policies, either expressed or implied, of the U.S. Government. The U.S. Government is authorized to reproduce and distribute reprints for Government purposes notwithstanding any copyright notation here on.

{\small
\bibliographystyle{ieee_fullname}
\bibliography{main_bib}

\begin{thebibliography}{10}\itemsep=-1pt

\bibitem{chao2018learning}
Yu-Wei Chao, Yunfan Liu, Xieyang Liu, Huayi Zeng, and Jia Deng.
\newblock Learning to detect human-object interactions.
\newblock In {\em 2018 IEEE Winter Conference on Applications of Computer
  Vision (WACV)}, pages 381--389. IEEE, 2018.

\bibitem{chen2018knowledge}
Kan Chen, Jiyang Gao, and Ram Nevatia.
\newblock Knowledge aided consistency for weakly supervised phrase grounding.
\newblock In {\em CVPR}, 2018.

\bibitem{chen2018msrc}
Kan Chen, Rama Kovvuri, Jiyang Gao, and Ram Nevatia.
\newblock Msrc: Multimodal spatial regression with semantic context for phrase
  grounding.
\newblock {\em International Journal of Multimedia Information Retrieval},
  7(1):17--28, 2018.

\bibitem{chen2017query}
Kan Chen, Rama Kovvuri, and Ram Nevatia.
\newblock Query-guided regression network with context policy for phrase
  grounding.
\newblock In {\em Proceedings of the IEEE International Conference on Computer
  Vision}, pages 824--832, 2017.

\bibitem{chen2019knowledge}
Tianshui Chen, Weihao Yu, Riquan Chen, and Liang Lin.
\newblock Knowledge-embedded routing network for scene graph generation.
\newblock In {\em Proceedings of the IEEE Conference on Computer Vision and
  Pattern Recognition}, pages 6163--6171, 2019.

\bibitem{chen2017implementation}
Xinlei Chen and Abhinav Gupta.
\newblock An implementation of faster rcnn with study for region sampling.
\newblock {\em arXiv preprint arXiv:1702.02138}, 2017.

\bibitem{dai2017detecting}
Bo Dai, Yuqi Zhang, and Dahua Lin.
\newblock Detecting visual relationships with deep relational networks.
\newblock In {\em Proceedings of the IEEE Conference on Computer Vision and
  Pattern Recognition}, pages 3076--3086, 2017.

\bibitem{deng2009imagenet}
Jia Deng, Wei Dong, Richard Socher, Li-Jia Li, Kai Li, and Li Fei-Fei.
\newblock Imagenet: A large-scale hierarchical image database.
\newblock In {\em 2009 IEEE Conference on Computer Vision and Pattern
  Recognition}, pages 248--255. IEEE, 2009.

\bibitem{galleguillos2008object}
Carolina Galleguillos, Andrew Rabinovich, and Serge Belongie.
\newblock Object categorization using co-occurrence, location and appearance.
\newblock In {\em 2008 IEEE Conference on Computer Vision and Pattern
  Recognition}, pages 1--8. IEEE, 2008.

\bibitem{gao2018ican}
Chen Gao, Yuliang Zou, and Jia-Bin Huang.
\newblock ican: Instance-centric attention network for human-object interaction
  detection.
\newblock {\em arXiv preprint arXiv:1808.10437}, 2018.

\bibitem{gkioxari2018detecting}
Georgia Gkioxari, Ross Girshick, Piotr Doll{\'a}r, and Kaiming He.
\newblock Detecting and recognizing human-object interactions.
\newblock In {\em Proceedings of the IEEE Conference on Computer Vision and
  Pattern Recognition}, pages 8359--8367, 2018.

\bibitem{he2016deep}
Kaiming He, Xiangyu Zhang, Shaoqing Ren, and Jian Sun.
\newblock Deep residual learning for image recognition.
\newblock In {\em Proceedings of the IEEE conference on computer vision and
  pattern recognition}, pages 770--778, 2016.

\bibitem{hu2017modeling}
Ronghang Hu, Marcus Rohrbach, Jacob Andreas, Trevor Darrell, and Kate Saenko.
\newblock Modeling relationships in referential expressions with compositional
  modular networks.
\newblock In {\em Proceedings of the IEEE Conference on Computer Vision and
  Pattern Recognition}, pages 1115--1124, 2017.

\bibitem{hu2016natural}
Ronghang Hu, Huazhe Xu, Marcus Rohrbach, Jiashi Feng, Kate Saenko, and Trevor
  Darrell.
\newblock Natural language object retrieval.
\newblock In {\em Proceedings of the IEEE Conference on Computer Vision and
  Pattern Recognition}, pages 4555--4564, 2016.

\bibitem{kingma2014adam}
Diederik~P Kingma and Jimmy Ba.
\newblock Adam: A method for stochastic optimization.
\newblock {\em arXiv preprint arXiv:1412.6980}, 2014.

\bibitem{krishna2018referring}
Ranjay Krishna, Ines Chami, Michael Bernstein, and Li Fei-Fei.
\newblock Referring relationships.
\newblock In {\em Proceedings of the IEEE Conference on Computer Vision and
  Pattern Recognition}, pages 6867--6876, 2018.

\bibitem{krishna2017genome}
Ranjay Krishna, Yuke Zhu, Oliver Groth, Justin Johnson, Kenji Hata, Joshua
  Kravitz, Stephanie Chen, Yannis Kalantidis, Li-Jia Li, David~A Shamma, et~al.
\newblock Visual genome: Connecting language and vision using crowdsourced
  dense image annotations.
\newblock {\em International Journal of Computer Vision}, 123(1):32--73, 2017.

\bibitem{laberge1997shifting}
David LaBerge, Robert~L Carlson, John~K Williams, and Blynn~G Bunney.
\newblock Shifting attention in visual space: Tests of moving-spotlight models
  versus an activity-distribution model.
\newblock {\em Journal of Experimental Psychology: Human Perception and
  Performance}, 23(5):1380, 1997.

\bibitem{li2017vip}
Yikang Li, Wanli Ouyang, Xiaogang Wang, and Xiao'ou Tang.
\newblock Vip-cnn: Visual phrase guided convolutional neural network.
\newblock In {\em Proceedings of the IEEE Conference on Computer Vision and
  Pattern Recognition}, pages 1347--1356, 2017.

\bibitem{yikangfactorizable}
Yikang Li, Wanli Ouyang, Bolei Zhou, Jianping Shi, Chao Zhang, and Xiaogang
  Wang.
\newblock Factorizable net: an efficient subgraph-based framework for scene
  graph generation.
\newblock In {\em Proceedings of the European Conference on Computer Vision
  (ECCV)}, pages 335--351, 2018.

\bibitem{li2017scene}
Yikang Li, Wanli Ouyang, Bolei Zhou, Kun Wang, and Xiaogang Wang.
\newblock Scene graph generation from objects, phrases and region captions.
\newblock In {\em Proceedings of the IEEE International Conference on Computer
  Vision}, pages 1261--1270, 2017.

\bibitem{li2019transferable}
Yong-Lu Li, Siyuan Zhou, Xijie Huang, Liang Xu, Ze Ma, Hao-Shu Fang, Yanfeng
  Wang, and Cewu Lu.
\newblock Transferable interactiveness knowledge for human-object interaction
  detection.
\newblock In {\em Proceedings of the IEEE Conference on Computer Vision and
  Pattern Recognition}, pages 3585--3594, 2019.

\bibitem{liang2018visual}
Kongming Liang, Yuhong Guo, Hong Chang, and Xilin Chen.
\newblock Visual relationship detection with deep structural ranking.
\newblock In {\em Thirty-Second AAAI Conference on Artificial Intelligence},
  2018.

\bibitem{liu2019knowledge}
Xuejing Liu, Liang Li, Shuhui Wang, Zheng-Jun Zha, Li Su, and Qingming Huang.
\newblock Knowledge-guided pairwise reconstruction network for weakly
  supervised referring expression grounding.
\newblock In {\em Proceedings of the 27th ACM International Conference on
  Multimedia}, pages 539--547. ACM, 2019.

\bibitem{lu2016visual}
Cewu Lu, Ranjay Krishna, Michael Bernstein, and Li Fei-Fei.
\newblock Visual relationship detection with language priors.
\newblock In {\em European Conference on Computer Vision}, pages 852--869.
  Springer, 2016.

\bibitem{pennington2014glove}
Jeffrey Pennington, Richard Socher, and Christopher~D. Manning.
\newblock Glove: Global vectors for word representation.
\newblock In {\em Empirical Methods in Natural Language Processing (EMNLP)},
  pages 1532--1543, 2014.

\bibitem{peyre2019detecting}
Julia Peyre, Ivan Laptev, Cordelia Schmid, and Josef Sivic.
\newblock Detecting unseen visual relations using analogies.
\newblock In {\em Proceedings of the IEEE International Conference on Computer
  Vision}, pages 1981--1990, 2019.

\bibitem{Plummer2017PhraseLA}
Bryan~A Plummer, Arun Mallya, Christopher~M Cervantes, Julia Hockenmaier, and
  Svetlana Lazebnik.
\newblock Phrase localization and visual relationship detection with
  comprehensive image-language cues.
\newblock In {\em Proceedings of the IEEE International Conference on Computer
  Vision}, pages 1928--1937, 2017.

\bibitem{qi2018learning}
Siyuan Qi, Wenguan Wang, Baoxiong Jia, Jianbing Shen, and Song-Chun Zhu.
\newblock Learning human-object interactions by graph parsing neural networks.
\newblock In {\em Proceedings of the European Conference on Computer Vision
  (ECCV)}, pages 401--417, 2018.

\bibitem{ren2015faster}
Shaoqing Ren, Kaiming He, Ross Girshick, and Jian Sun.
\newblock Faster r-cnn: Towards real-time object detection with region proposal
  networks.
\newblock In {\em Advances in neural information processing systems}, pages
  91--99, 2015.

\bibitem{rohrbach2016grounding}
Anna Rohrbach, Marcus Rohrbach, Ronghang Hu, Trevor Darrell, and Bernt Schiele.
\newblock Grounding of textual phrases in images by reconstruction.
\newblock In {\em European Conference on Computer Vision}, pages 817--834.
  Springer, 2016.

\bibitem{sadhu2019zero}
Arka Sadhu, Kan Chen, and Ram Nevatia.
\newblock Zero-shot grounding of objects from natural language queries.
\newblock In {\em Proceedings of the IEEE International Conference on Computer
  Vision}, pages 4694--4703, 2019.

\bibitem{simonyan2014very}
Karen Simonyan and Andrew Zisserman.
\newblock Very deep convolutional networks for large-scale image recognition.
\newblock {\em arXiv preprint arXiv:1409.1556}, 2014.

\bibitem{wan2019pose}
Bo Wan, Desen Zhou, Yongfei Liu, Rongjie Li, and Xuming He.
\newblock Pose-aware multi-level feature network for human object interaction
  detection.
\newblock In {\em Proceedings of the IEEE International Conference on Computer
  Vision}, pages 9469--9478, 2019.

\bibitem{wang2016structured}
Mingzhe Wang, Mahmoud Azab, Noriyuki Kojima, Rada Mihalcea, and Jia Deng.
\newblock Structured matching for phrase localization.
\newblock In {\em European Conference on Computer Vision}, pages 696--711.
  Springer, 2016.

\bibitem{xu2019learning}
Bingjie Xu, Yongkang Wong, Junnan Li, Qi Zhao, and Mohan~S Kankanhalli.
\newblock Learning to detect human-object interactions with knowledge.
\newblock In {\em Proceedings of the IEEE Conference on Computer Vision and
  Pattern Recognition}, 2019.

\bibitem{xu2017scenegraph}
Danfei Xu, Yuke Zhu, Christopher Choy, and Li Fei-Fei.
\newblock Scene graph generation by iterative message passing.
\newblock In {\em Proceedings of the IEEE Conference on Computer Vision and
  Pattern Recognition}, pages 5410--5419, 2017.

\bibitem{Yang_2018_ECCV}
Jianwei Yang, Jiasen Lu, Stefan Lee, Dhruv Batra, and Devi Parikh.
\newblock Graph r-cnn for scene graph generation.
\newblock In {\em Proceedings of the European Conference on Computer Vision
  (ECCV)}, pages 670--685, 2018.

\bibitem{yang2018shuffle}
Xu Yang, Hanwang Zhang, and Jianfei Cai.
\newblock Shuffle-then-assemble: Learning object-agnostic visual relationship
  features.
\newblock In {\em Proceedings of the European Conference on Computer Vision
  (ECCV)}, pages 36--52, 2018.

\bibitem{yin2018zoom}
Guojun Yin, Lu Sheng, Bin Liu, Nenghai Yu, Xiaogang Wang, Jing Shao, and Chen
  Change~Loy.
\newblock Zoom-net: Mining deep feature interactions for visual relationship
  recognition.
\newblock In {\em European Conference on Computer Vision}, pages 322--338,
  2018.

\bibitem{yu2018mattnet}
Licheng Yu, Zhe Lin, Xiaohui Shen, Jimei Yang, Xin Lu, Mohit Bansal, and
  Tamara~L Berg.
\newblock Mattnet: Modular attention network for referring expression
  comprehension.
\newblock In {\em Proceedings of the IEEE Conference on Computer Vision and
  Pattern Recognition}, pages 1307--1315, 2018.

\bibitem{yu2017visual}
Ruichi Yu, Ang Li, Vlad~I Morariu, and Larry~S Davis.
\newblock Visual relationship detection with internal and external linguistic
  knowledge distillation.
\newblock In {\em Proceedings of the IEEE International Conference on Computer
  Vision}, pages 1974--1982, 2017.

\bibitem{zellers2018motif}
Rowan Zellers, Mark Yatskar, Sam Thomson, and Yejin Choi.
\newblock Neural motifs: Scene graph parsing with global context.
\newblock In {\em Proceedings of the IEEE Conference on Computer Vision and
  Pattern Recognition}, pages 5831--5840, 2018.

\end{thebibliography}
}

\end{document}